\documentclass[10pt, conference, a4paper]{IEEEtran}
\ifCLASSINFOpdf
  \usepackage[pdftex]{graphicx}
  \usepackage{pifont}
\else

\fi

\usepackage{amsmath}
\usepackage{courier}
\usepackage[bookmarks=false]{hyperref}
\usepackage{multirow}

\DeclareMathOperator*{\argmax}{argmax} 

\begin{document}

\textsc{\large \bf IEEE Copyright Notice}\\

\textsuperscript{\textcopyright} 2018 IEEE. Personal use of this material is permitted. Permission from IEEE must be obtained for all other uses, in any current or future media, including reprinting/republishing this material for advertising or promotional purposes, creating new collective works, for resale or redistribution to servers or lists, or reuse of any copyrighted component of this work in other works. \\

E. Razinkov, I. Saveleva and J. Matas, "ALFA: Agglomerative Late Fusion Algorithm for Object Detection," 2018 24th International Conference on Pattern Recognition (ICPR), Beijing, 2018, pp. 2594-2599.\\
doi: 10.1109/ICPR.2018.8545182\\
keywords: {convolution;feedforward neural nets;image fusion;object detection;pattern clustering;ALFA;agglomerative late Fusion algorithm;object detection;agglomerative clustering;object detector predictions;bounding box locations;weighted combination;PASCAL VOC 2007;PASCAL VOC 2012;dynamic belief fusion;single object hypothesis;bounding boxes clustering;SSD;DeNet;Faster R-CNN;baseline combination strategies;DBF;Detectors;Proposals;Object detection;Feature extraction;Convolutional codes;Prediction algorithms;Heuristic algorithms}.\\
URL: \url{http://ieeexplore.ieee.org/stamp/stamp.jsp?tp=&arnumber=8545182&isnumber=8545020}

%
\title{ALFA: Agglomerative Late Fusion Algorithm \\ for Object Detection}


\author{\IEEEauthorblockN{Evgenii Razinkov}
\IEEEauthorblockA{Institute of Computational Mathematics\\and Information Technologies\\
Kazan Federal University, Russia\\
Email: Evgenij.Razinkov@kpfu.ru}
\and

\IEEEauthorblockN{Iuliia Saveleva}
\IEEEauthorblockA{Institute of Computational Mathematics\\and Information Technologies\\
Kazan Federal University, Russia\\
Email: JuOSaveleva@stud.kpfu.ru}
\and
\IEEEauthorblockN{Ji\v{r}i Matas}
\IEEEauthorblockA{Faculty of Electrical Engineering\\Czech Technical University in Prague \\
Prague, Czech Republic}
Email: matas@cmp.felk.cvut.cz
}

\maketitle

\begin{abstract}

We propose ALFA -- a novel late fusion algorithm for object detection. ALFA is based on agglomerative clustering of object detector predictions taking into consideration both the bounding box locations and the class scores. Each cluster represents a single object hypothesis whose location is a weighted combination of the clustered bounding boxes. 

ALFA was evaluated using combinations of a pair (SSD and DeNet) and a triplet (SSD, DeNet and Faster R-CNN) of recent
object detectors that are close to the state-of-the-art. ALFA achieves state of the art results on PASCAL VOC 2007 and PASCAL VOC 2012, outperforming the individual detectors as well as baseline combination strategies,  achieving up to 32\% lower error than the best individual detectors and up to 6\% lower error than the reference fusion algorithm DBF -- Dynamic Belief Fusion.
\end{abstract}


\IEEEpeerreviewmaketitle

\section{Introduction}

Object detection is an important and challenging task in computer vision with a lot of applications. In recent years accuracy of object detectors significantly improved due to use of learning. R-CNN~\cite{rcnn} was the first breakthrough associated with using deep convolutional neural networks for object detection. Fast R-CNN~\cite{fastrcnn} and Faster R-CNN~\cite{fasterrcnn} develop the idea further improving both detection speed and accuracy. You Only Look Once~\cite{yolo}, Single-Shot Detector~\cite{ssd} introduced more lightweight approach and end-to-end training achieving remarkable accuracy while operating in real-time. YOLOv2~\cite{yolo2} and DSSD~\cite{dssd} and most recent DeNet~\cite{denet} object detector push these boundaries even further.

It is known that aggregating machine learning algorithms in an ensemble tend to improve performance~\cite{ensemble}. Aggregating different object detectors usually called \emph{fusion} was shown to improve accuracy as well. So called \emph{late fusion} methods treat base object detectors as black boxes using only their predictions as input.

This paper studies the problem of late fusion on object detector outputs. We want to explore whether modern deep object detectors differ from each other enough so their fusion could show significant performance boost in comparison to individual detectors. 
\begin{figure}[t]
  \includegraphics[width=\columnwidth]{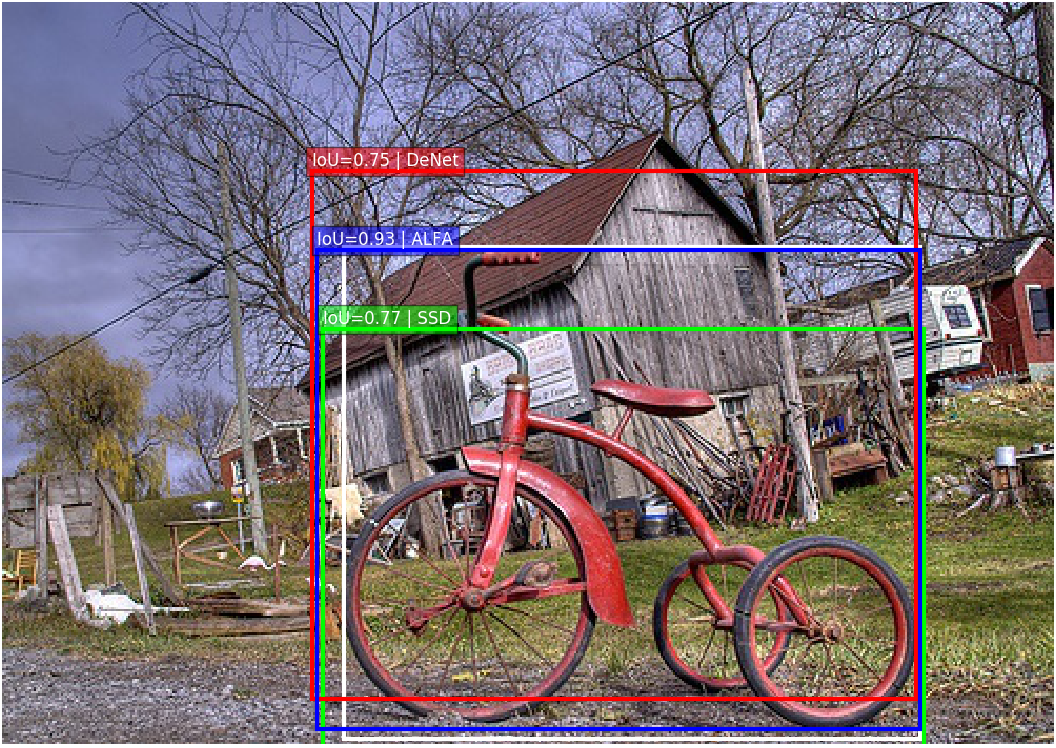}
  \caption{Image from PASCAL VOC 2007 \texttt{test} set. Bounding boxes and IoU with ground truth: DeNet -- red (IoU = 0.75); SSD -- green (IoU = 0.77); ALFA -- blue (IoU = 0.93). Ground truth bounding box is in white.}
  \label{bike_pic}
\end{figure}

The problem of object detector fusion has been addressed in the literature. Detect2Rank~\cite{detect2rank} uses Learn2Rank algorithms to rank detections from different object detectors. This goal is achieved by representing each detection with a feature vector and learning a ranking system on a validation set. Handcrafted feature vector includes information about detector-detector context, object saliency and object-object relation information. Ranking is learned using L2 regularized support vector classifier, logistic regressor, support vector regressor and RankSVM. Non-maximum suppression is then used to remove multiple detections of the same object with lower scores. 

Dynamic Belief Fusion is a late fusion algorithm that re-scores detections confidence using Dempster-Shafer theory~\cite{dbf}. DBF solves object detection as a binary classification problem -- each bounding box either contains an object or not. Using precision-recall curves built for each base detector and for abstract hyperparameter-dependent ``best possible detectors'' all detections are re-scored and non-maximum suppression is applied. Authors of~\cite{dbf} claim that DBF outperforms all existing fusion methods including Detect2Rank. Unfortunately, the paper does not go into much detail regarding extending binary DBF framework to multiclass object detection scenario. 

Faster R-CNN with ResNet-101 as a base network currently acheive state of the art results in object detection in terms of mAP while being quite slow. These object detectors were used as base detectors in an ensemble in~\cite{resnet}. Several Region Proposal Networks and Classification Networks were trained independently, at test time union set of region predictions from all RPNs is classified by an ensemble of classification networks. Non-maximum suppression is run afterwards. Using three models in an ensemble boosts performance from 34.9 mAP to 37.4 mAP on MS COCO object detection dataset~\cite{mscoco}.  

None of these methods provides new bounding boxes as fusion outputs, bounding box for the object is always one of the bounding boxes predicted by base detectors. We believe that combining bounding boxes from all base detectors can lead to better object localization. 

\subsection{Contribution}

Our contribution is as follows:
\begin{itemize}
\item We show that modern deep object detectors differ enough so their combination significantly outperforms individual detectors.
\item We propose novel Agglomerative Late Fusion Algorithm~(ALFA) for object detection that shows state of the art results on PASCAL VOC 2007~\cite{pascalvoc} and PASCAL VOC 2012~\cite{pascalvoc2012} object detection datasets reducing error by up to 32\% in comparison with individual detectors and by up to 6\% in comparison with the reference fusion methods. 
\item We clarify DBF extension to multiclass object detection scenario and provide experimental results for modern detectors fusion using DBF.
\item We make source code for ALFA and our implementation of DBF publicly available making results of our research reproducible: \url{http://github.com/IuliiaSaveleva/ALFA}

\end{itemize}

\section{Base Object Detectors}

Assume $N$ base object detectors $D_1, D_2, ..., D_N$ for $K$ classes. After processing image $I$, detector $i$ outputs $m_i$ predictions of object presence. Each prediction consists of bounding box coordinates and a $(K+1)$-tuple of class scores:
\begin{equation}
D_i(I) = \left\{ (r_1, c_1), ..., (r_{m_i}, c_{m_i})  \right\}, \quad i = 1, ..., N,
\end{equation}
where $r_j$ are the four coordinates of the axis-aligned bounding box and $c_j$ are the class scores for the $j$-th detected object. 
$$
r = \left( x_{tl}, y_{tl}, x_{br}, y_{br} \right),  
$$
where $(x_{tl}, y_{tl})$ are the coordinates of the top-left corner of the bounding box and $(x_{br}, y_{br})$ are the coordinates of the bottom-right corner of the bounding box and
\begin{align}
0 \le x_{tl} < x_{br} < I_{width}, \quad
0 \le y_{tl} < y_{br} < I_{height}. \nonumber
\end{align}
Class scores $c_j$ is a tuple
$$
c_j = \left(c_j^{(0)}, c_j^{(1)}, ..., c_j^{(K)} \right), 
$$
where $c_j^{(0)}$ is  the ``no object'' probability and $c_j^{(1)}, c_j^{(2)}, ..., c_j^{(K)}$ are the probabilities for $K$ classes, 
$$
\sum_{k = 0}^K c^{(k)} = 1, \quad c^{(k)} \ge 0, \quad k = 0, ..., K.
$$

The core problem for fusion of multiple detectors is the decision which detections are due to the same object. 

\section{The Proposed Method} 

We formulate the problem of late fusion as agglomerative clustering and  propose a parametrized similarity function that takes into account both spatial properties of the two predictions and their class scores. Parameters are learned on validation set. We define object proposal to be the set of predictions forming a cluster. A proposal therefore comprises one or more predictions. 

Assuming that predictions with similar bounding boxes and class scores often correspond to the same object, only one bounding box and class score tuple is output per proposal.
We explore several strategies for estimating the object proposal bounding box and classification. 
We employ non-maximum suppression with IoU threshold 0.5 to remove remaining multiple detections of the same object.

\subsection{The Clustering Method}
To define clustering procedure one needs to define similarity score function between two samples, similarity score functions between two clusters and stopping criteria.

\subsubsection{Similarity Between Proposals}
We assume that all predictions associated with the object proposal are due to the same object. Hence, each prediction bounding box and class scores should be similar to each other prediction bounding box and class scores. This intuition strongly suggests using complete-link clustering. Let $C_i$ and $C_j$ be two clusters and $\sigma(p, \tilde{p})$ -- similarity score function between predictions $p$ and $\tilde{p}$ that will be defined later. We define the following similarity score function for prediction clusters:
\begin{equation}
\sigma(C_i, C_j) = \min_{p \in C_i, \tilde{p} \in C_j} \sigma(p, \tilde{p}).
\end{equation}

\subsubsection{Stopping Criteria}

Clustering stops when
\begin{equation}
\max_{i, j} \sigma(C_i, C_j) < \tau,
\end{equation}
where $\tau$ is a hyperparameter. Therefore, every two predictions associated with the same object proposal have similarity score no less than $\tau$.

\subsubsection{Similarity Between Predictions}
We assume that detections with substantially different class scores often correspond to different objects on the image. Therefore, we want to incorporate class scores similarity into our similarity metric.
We employ Bhattacharyya coefficient as a measure of similarity between class scores:
\begin{equation}
BC(\bar{c}_i, \bar{c}_j) = \sum_{k = 1}^K \sqrt{\bar{c}_i^{(k)}\bar{c}_j^{(k)}},
\end{equation}
where $\bar{c}$ is obtained from class score tuple $c$ by omitting the zeroth \emph{``no object''} component and renormalizing:
$$
\bar{c}^{(k)} = \frac{c^{(k)}}{1 - c^{(0)}}, \quad k = 1, ... K.
$$

Note that $BC(\bar{c}_i, \bar{c}_j) \in [0,1]$ and equals 1 if and only if $\bar{c}_i = \bar{c}_j$. 

Two detections with similar shape, size and position on the image will often correspond to the same object. To decide whether two detections correspond to the same object we should account for similarity between their bounding boxes.
We use intersection over union coefficient which is widely used as a measure of similarity between bounding boxes:
\begin{equation}
IoU(r_i, r_j) = \frac{r_i \cap r_j}{r_i \cup r_j}.
\end{equation}

We propose the following measure of similarity between object detectors predictions $p_i = (r_i, c_i)$ and $p_j = (r_j, c_j)$ that takes into account both class scores similarity and similarity of bounding boxes:
\begin{equation}
\sigma(p_i, p_j) = IoU(r_i, r_j)^\gamma\cdot BC(\bar{c}_i, \bar{c}_j) ^{1 - \gamma},
\end{equation}
where $\gamma \in [0, 1]$ is a hyperparameter. Note that $\sigma(p_i, p_j) \in [0, 1]$, $\sigma(p_i, p_j)$ equals 0 if bounding boxes for predictions $p_i$ and $p_j$ do not intersect or their class scores do not overlap. With $\gamma \in (0, 1)$, $\sigma(p_i, p_j) = 1$ if and only if bounding boxes and class scores for $p_i$ and $p_j$ are equal correspondingly.

Since NMS is applied to every base detector predictions before the fusion procedure, we assume that multiple similar detections from the same base detector are often due to different objects. Therefore, detections from the same base detector should be assigned to different clusters. To achieve that, we set similarity scores between predictions from the same base detector to zero.

\subsection{Decision on the class of the cluster}

Assume predictions from detectors $D_{i_1}, D_{i_2}, ..., D_{i_s}$ were assigned to object proposal $\pi$. 
Confidence of the prediction is a score value of the predicted class.
Object detector fusion is not straightforward because of variable number of predictions corresponding to each object proposal.

To account for missing detections associated with an object proposal we assign an additional low-confidence class scores tuple to this object proposal for every detector that missed. Low-confidence class scores should not influence class prediction for the object proposal, therefore, last $K$ components of a tuple should be equal:
$$
c_{lc}^{(i)} = c_{lc}^{(j)}, \quad i, j = 1, ..., K.
$$
Moreover, since recall of modern detectors is less than 1, missed detection does not always mean that there is no object. We use the following low-confidence class scores:
\begin{equation}
c_{lc} = \left(1 - \varepsilon, \frac{\varepsilon}{K}, \frac{\varepsilon}{K}, ..., \frac{\varepsilon}{K}   \right),
\end{equation}
where $\varepsilon$ is a hyperparameter. Assigning low-confidence class scores to object proposal significantly lowers the confidence of the resulting predictions if $\varepsilon$ is close to zero. 

Possible strategies for dealing with class scores aggregation are inspired by classification ensembles and include:
\begin{itemize}
\item Choosing most confident prediction associated with the object proposal:
\begin{equation}
c_{\pi} = c_j,
\end{equation}
where 
\begin{equation}
j = \argmax_{i \in \pi} \max_{1 \le k \le K} c_i^{(k)}.
\end{equation}
\item \emph{Averaging fusion.} Averaging class scores vectors of associated predictions:
\begin{equation}
c_{\pi}^{(k)} = \frac{1}{N} \left( \sum_{d = 1}^s c_{i_d}^{(k)} +  (N - s)\cdot c_{lc}^{(k)}  \right), k = 0, ..., K.
\end{equation}
\item \emph{Multiplication fusion.} Multiplying class scores corresponding to the same class and renormalizing:
\begin{equation}
c_{\pi}^{(k)} =  \frac{\tilde{c}_{\pi}^{(k)}}{\sum_{i} \tilde{c}_{\pi}^{(i)}},
\end{equation}
where
\begin{equation}
\tilde{c}_{\pi}^{(k)} = \left(c_{lc}^{(k)} \right)^{N - s} \prod_{d = 1}^s c_{i_d}^{(k)}, \quad k = 0, ..., K.
\end{equation}
\end{itemize}

\subsection{Object localization}

We have explored several object localization strategies:
\begin{itemize}
\item \emph{Bounding box associated with the most confident prediction.} Choosing bounding box corresponding to the most confident prediction associated with the object proposal:
\begin{equation}
r_{\pi} = r_j,
\end{equation}
where 
\begin{equation}
j = \argmax_{i \in \pi} \max_{1 \le k \le K} c_i^{(k)}.
\end{equation}
This principle is usually used in non-maximum suppression algorithms. NMS is often applied as a final step in fusion algorithms.
\item \emph{Averaged bounding box.} Averaging outputs of the weak learners is a common way to aggregate predictions to produce output of an ensemble:
\begin{equation}
r_{\pi} = \frac{1}{|\pi|}\sum_{i \in \pi} r_{i}.
\end{equation}
\item \emph{Bounding box obtained by weighted averaging.} Correct object localization usually associated with more confident prediction. Following this intuition we assign weights to bounding boxes based on prediction confidence.
Fusion outputs the following bounding box for object proposal~$\pi$:
\begin{equation}
r_{\pi} = \sum_{i \in \pi} \frac{w_i}{\sum_{j \in \pi} w_j}\cdot r_{i},
\end{equation}
where
\begin{equation}
w_i = \max_{1 \le k \le K}(c_{i}^{(k)}).
\end{equation}
\item \emph{Average weighted by proposal class confidence} This localization strategy is similar to the previous one except we use class scores of the label that is chosen for the object proposal as weights:
\begin{equation}
r_{\pi} = \frac{1}{\sum_{i \in \pi} c_{i}^{(l)}} \sum_{i \in \pi} c_{i}^{(l)} \cdot r_{i},
\end{equation}
where
\begin{equation}
l = \argmax_{k \ge 1} c_{\pi}^{(k)}.
\end{equation}
This strategy relies on the assumption that localization of an object depends on the predicted class scores.
\end{itemize}

\section{Experiments}

We evaluated ALFA and baseline methods on PASCAL VOC 2007 and VOC 2012 object detection datasets using the following base object detectors:
\begin{itemize}
\item Single Shot Detector: SSD-300 with VGG-16 as a base network trained on PASCAL VOC 07+12 \texttt{trainval} dataset, i.e. on the union of the training and validation images of the PASCAL VOC 2007 and VOC 2012 datasets.
\item DeNet with \emph{skip} layers and ResNet-101 as a base network trained on PASCAL VOC 07+12 \texttt{trainval} dataset.
\item Faster R-CNN with ResNet-101 as a base network trained on PASCAL VOC 07+12 \texttt{trainval} dataset. We do not use global context, multi-scale testing and bounding box refinement introduced in~\cite{resnet}.
\end{itemize}
These object detection methods were selected for three reasons. First, none of them is slow and its combination that requires running all three is not impractical. Second, these are commonly used object detectors with performance no far from the state of the art. Third, for SSD and DeNet the source code and the weights for the object detector are available from the authors.

\subsection{Base Detectors}

Faster R-CNN~\cite{fasterrcnn} uses Region Proposal Network which takes last feature map as an input and outputs a number of bounding boxes that could potentially contain objects. The second component -- Fast R-CNN -- is used to determine whether patricular proposed bounding box contains an object of certain class or not. While in the original paper VGG~\cite{vgg} is used as base network, it was shown in~\cite{resnet} that using ResNet-101 as base network improves accuracy significantly. Faster R-CNN with ResNet-101 base network with multi-scale testing, inclusion of context and bounding box refinement, sometimes referred to as Faster R-CNN+++, shows state of the art performance in terms of accuracy but is slower than YOLO, SSD and DeNet.

Single Shot Detector~\cite{ssd} does not use RPN but instead relies on convolutional detectors to classify objects on different scales. SSD uses VGG convolutional network with additional convolutional layers for feature extraction. Convolutional detectors for different feature maps predict object classes and offsets with respect to corresponding default bounding boxes. Making use of feature maps at different depths allows SSD to detect objects with different sizes. Neurons from not-too-deep layers of CNN have smaller receptive fields thus are suitable for small objects detection. Neurons at deeper layers have larger receptive fields that allow detection of bigger objects. 
Convolutional detector consists of $(K + 1) + 4$ convolutions with $3 \times 3 \times d$ filter size predicting class scores for $K$ classes along with ''no object'' class and offsets of the bounding box with respect to associated with the detector default bounding box. Non-maximum suppression is applied to all the predictions from convolutional detectors at each scale.

DeNet~\cite{denet} relies on region-of-interest estimation based on corner detection. DeNet predicts for each image pixel how likely it is a certain corner of an object bounding box. Predicted corners form regions-of-interest that are classified by neural network. Therefore, DeNet uses neither Region Proposal Network nor handcrafted default boxes for object localization. ResNet-101~\cite{resnet} is used as a base network, deconvolutions are applied to the last layer of ResNet in order to increase localization accuracy. DeNet acheives remarkable performance in terms of speed and accuracy outperforming SSD. 

\subsection{Baseline methods}

\subsubsection{Dynamic Belief Fusion}
We use Dynamic Belief Fusion as a baseline since it was shown to outperform other fusion methods~\cite{dbf}. We use our own implementation of DBF since source code for DBF is not available. 

Dynamic Belief Fusion~\cite{dbf} relies on precision-recall curves, thus, binary decision problem is assumed. While~\cite{dbf} clearly states that DBF is applied to object detection on multiclass PASCAL VOC 2007 dataset, only binary procedure is described in the paper and it is not clear how to extend DBF to multiclass scenario. We tested three options how to handle multiclass case in DBF:
\begin{enumerate}
\item Object detection is treated as a binary problem -- object is either present in a bounding box or not. In this case each detector has one class-agnostic precision-recall curve. We do not aggregate bounding boxes with different labels so decision on a final label is trivial.
\item Each base detector is considered as $K$ class-specific binary object detectors and every detection with class scores $c$ is treated as $K$ detections with confidence values $c^{(1)}, ...., c^{(K)}$. Each base detector has $K$ class-specific precision-recall curves associated with it. We also assume $K$ class-specific ``best possible detectors''. 
\item Each base object detector is again treated as $K$ class-specific binary object detectors with $K$ class-specific precision-recall curves but this time there is only one class-agnostic ``best possible detector''.
\end{enumerate}
We have implemented all three versions and measured their performance. We use the third variant as a baseline since it outperforms the first and the second option. 

\subsubsection{Non-Maximum Suppression}

We use Greedy Non-Maximum Suppression as our second baseline since it was successfully used in~\cite{resnet} to aggregate object detectors outputs. GreedyNMS suppresses every prediction that overlaps with $\mbox{IoU} > 0.5$ with more confident prediction of the same class.

\subsection{Note on confidence thresholds}

It is possible to set confidence threshold of any base detector to exclude least confident detections. 

We can set threshold $\theta$ for an object detector and consider only those predictions with class scores $c$ for which the following holds:
$$
c^{(l)} \ge \theta,
$$
where $l$ is the predicted label.

Value of $\theta$ for base detectors affects fusion performance. Bigger $\theta$ results in decreased number of predictions speeding up fusion algorithm since amount of computation is proportional to the number of predictions.

We choose $\theta$ for DBF and NMS to maximize fusion mAP. For ALFA we use two thresholds chosen to achieve two different goals:
(i) maximize fusion mAP and (ii) maximize mAP while keeping fusion computation under 2-3 ms. We refer to this version as ``Fast ALFA'' in experiment results. 
Fusion performance is measured on Intel Core i5-6400.

For all tested fusion algorithms smaller values of $\theta$ result in higher mAP and lower speed performance. We use $\theta = 0.015$ for NMS, DBF and ALFA. We use  $\theta = 0.05$ for Fast ALFA. 


\subsection{Note on evaluation procedure}

Object detector outputs a class scores vector for every detection. There are two ways of handling obtained class scores vectors:
\begin{itemize}
\item Consider every detection with class scores vector $c$ and bounding box $r$  to be a single detection with label $l$, where $l = \argmax_{ 1\le k \le K} c^{(k)}$, confidence $c^{(l)}$ and bounding box $r$. We refer to mean average precision computed this way as {\it mAP-s}.
\item Treat every detection with class scores vector $c$ and bounding box $r$ as multiple detections sharing the same bounding box -- one detection for each label $l$, where $l = 1, ..., K$, with confidence value $c^{(l)}$ and bounding box $r$. 

Since this approach is more common it is referred to as {\it mAP} in experiments results. 
\end{itemize}



\subsection{Results on PASCAL VOC 2007}

We employ 5-fold cross-validation on PASCAL VOC 2007 \texttt{test} to choose optimal lozalization and recognition strategies and to adjust hyperparameters $\gamma$, $\tau$ and $\varepsilon$. Best performing localization strategy is averaging weighted by proposal class confidence. Best hyperparameter values obtained through cross-validation are listed in Table~\ref{parameters_table}. 

Precision-recall curves for DBF are also estimated using 5-fold cross-validation.

During 5-fold cross-validation mean average precision is computed as follows:
\begin{itemize}
\item We compute average precision $AP_i^{(k)}$ on the $i$-th test part, $i = 1, ..., 5$, for each class $k$, $k = 1, ..., K$.
\item We compute weights for each class for the $i$-th test part:
$$
\alpha_i^{(k)} = {n_i^{(k)}}/{n^{(k)}},
$$
where $n_i^{(k)}$ is the number of objects of class $k$ in the $i$-th test part and $n^{(k)}$ is the number of objects of class $k$ in the whole test set.
\item Average precision for each class is computed as follows:
$$
AP^{(k)} = \sum_{i} \alpha_i^{(k)} AP_i^{(k)}.
$$
\item Mean average precision:
$$
\mbox{\it mAP} = \frac{1}{K} \sum_{k = 1}^K \mbox{\it AP}^{(k)}.
$$
\end{itemize}
This procedure introduces small bias. To make the comparison fair we use the same mAP computation procedure for all individual detectors and fusion methods on PASCAL VOC 2007.

Overall mAP and detection speed are provided in Table~\ref{map_fps}.

\subsection{Results on PASCAL VOC 2012}

We also evaluated all reviewed fusion methods performance on PASCAL VOC 2012 \texttt{test} set. Hyperparameters of ALFA and DBF were adjusted on PASCAL VOC 2007 \texttt{test} set. Results are presented in Table~\ref{map_fps}.
Surprisingly, DBF shows worse results than NMS in this scenario. We think that DBF does not generalize well across datasets with different difficulty since it relies on precision-recall curves.

\begin{table}[!h]
\renewcommand{\arraystretch}{1.3}
\caption{Best cross-validated hyperparameter values}
\label{parameters_table}
\centering
\begin{tabular}{|c|c|c|c|c|}
\hline
 & \multicolumn{2}{|c|}{ \begin{tabular}{@{}c@{}} Optimal values \\ (SSD + DeNet) \end{tabular} } & \multicolumn{2}{|c|}{ \begin{tabular}{@{}c@{}} Optimal values \\ (FRCNN + SSD + DeNet) \end{tabular}} \\
\hline
Evaluation & mAP-s & mAP & mAP-s & mAP \\

\hline
\begin{tabular}{@{}c@{}} Classification \\ confidence \end{tabular} & Multiplication & Averaging & Multiplication& Averaging\\
\hline
$\tau$ & 0.48 & 0.73 &  0.75 &  0.74\\
\hline
$\gamma$ & 0.22 & 0.25 & 0.28 & 0.30\\
\hline
$\varepsilon$ & 0.56 & 0.26 & 0.17 & 0.39 \\

\hline
\end{tabular}
\end{table}

\begin{table}[!h]
\renewcommand{\arraystretch}{1.3}
\caption{Results on PASCAL VOC 2007 and VOC 2012 }
\label{map_fps}
\centering
\begin{tabular}{|c|c|c|c|c|c|}
\hline
\multirow{3}{*}{Detector}& \multirow{3}{*}{\begin{tabular}{@{}c@{}} fps  \\ (Hz)\end{tabular}} & \multicolumn{2}{|c|}{PASCAL VOC 2007}  &  \multicolumn{2}{|c|}{PASCAL VOC 2012} \\
\cline{3-6}
 & & \begin{tabular}{@{}c@{}} mAP-s \\ (\%) \end{tabular}&\begin{tabular}{@{}c@{}} mAP \\ (\%) \end{tabular}& \begin{tabular}{@{}c@{}} mAP-s \\ (\%) \end{tabular}& \begin{tabular}{@{}c@{}} mAP \\ (\%) \end{tabular}\\
\hline
Faster R-CNN & 7  & 77.95 & 78.83 & 72.72 & 73.59  \\
\hline
SSD300 & 59  & 79.26 & 80.37 & 72.89 & 74.17\\
\hline
DeNet  & 33 & 78.09 & 79.26 &  70.73 & 72.10  \\
\hline
\hline
\multicolumn{6}{|c|}{SSD + DeNet}\\
\hline
NMS & 20.3  & 83.12 & 83.53 &  76.80 & 77.37\\
\hline
DBF & 16.9  & 83.29 & 83.88 &  75.74 & 76.38\\
\hline
Fast ALFA  &  \textbf{20.6}  & 83.87 & 84.32 & 76.97 & 77.82 \\
\hline
ALFA & 18.1  & \textbf{84.16} & \textbf{84.41} & \textbf{77.52} & \textbf{77.98}\\
\hline
\hline
\multicolumn{6}{|c|}{SSD + DeNet + Faster R-CNN}\\
\hline
NMS & \textbf{5.2}   & 84.31 & 84.43 &  78.11 & 78.34\\
\hline
DBF & 4.7  & 84.97 & 85.24 & 75.71 & 75.69\\
\hline 
Fast ALFA & \textbf{5.2} & 85.78 & 85.67 &   79.16 & 79.42\\
\hline
ALFA & 5.0  & \textbf{85.90} & \textbf{85.72} & \textbf{79.41} & \textbf{79.47}\\
\hline

\end{tabular}
\end{table}

\subsection{Ablation study}
The influence of the following design decisions is measured:
\begin{itemize}
\item Adding low-confidence class scores for every missed detection.
\item Aggregating class scores instead of using the most confident prediction. Adding low-confidence detections do not affect performance when using class scores for most confident prediction. 
\item Taking class scores into account while generating object proposals.
\item Using weighted average bounding box instead of using bounding box of the most confident prediction. 
\end{itemize}
Results of the ablation study are summarized in Table~\ref{ablation_study}.

\begin{table}[!h]
\renewcommand{\arraystretch}{1.3}

\caption{Effects of various parameters on fusion performance}
\label{ablation_study}
\centering
\resizebox{\columnwidth}{!}{%
\begin{tabular}{|c|c|c|c|c|c|c|c|}
\hline 
 & \multicolumn{7}{c|}{PASCAL VOC 2007 test}\\
\hline
\begin{tabular}{@{}c@{}} Adding low-confidence \\ class scores \end{tabular} &  &  & & \ding{51} &  \ding{51} & \ding{51} & \ding{51}\\
\hline
\begin{tabular}{@{}c@{}} Aggregating \\ class scores \end{tabular}& & \ding{51} &  & \ding{51} & \ding{51} & \ding{51} & \ding{51}\\
\hline
\begin{tabular}{@{}c@{}} Incorporate class scores \\ into distance metric \end{tabular} & & \ding{51} & \ding{51}&  & & \ding{51}  & \ding{51}\\
\hline
 \begin{tabular}{@{}c@{}} Aggregating\\bounding boxes\end{tabular} & & \ding{51} & \ding{51}&  &  \ding{51} &  & \ding{51}\\
\hline
\begin{tabular}{@{}c@{}} ALFA (FRCNN + \\ SSD + DeNet)\end{tabular} & 84.25 & 82.38 &  84.57 & 84.79  & 85.00 & 85.12 & 85.72\\
\hline
\end{tabular}
}
\end{table}

\section{Discussion}

ALFA shows higher mean average precision values for each detector combination on both PASCAL VOC 2007 and VOC 2012 when compared with base detectors and baseline fusion methods. Fast ALFA is slightly inferior to ALFA but still outperforming baseline fusion methods while being marginally faster.

To achieve close-to-real-time performance fusion method needs not only very fast base object detectors but also computationally light aggregation procedure. 
The key to fast implementation of ALFA is to break all objects into groups so that each object in the group is similar to at least one another object in this group. This step can be done quite effectively. Agglomerative clustering is then applied to each group independently. Computation time for our fusion procedure is as low as 1.2 ms for two detectors and 1.6 ms for three detectors (for $\theta = 0.05$) while code is written in Python. 

It is reasonable to assume, however, that ALFA performance will be slower for images crowded with objects from the same class with significantly overlapping bounding boxes.


\section{Conclusion}

We propose ALFA -- a novel late fusion algorithm for object detection. ALFA shows state of the art results on PASCAL VOC 2007 and VOC 2012 datasets outperforming individual detectors and existing fusion frameworks regardless of evaluation procedure while being computationally light. The classification error  expressed as $1 - \mbox{\it mAP}$, is reduced by up to 32\% in comparison to the base detectors and by up to 6\% when compared with the state of the art fusion method DBF, that, as experiments indicate, does not generalize well across different datasets.


\section*{Acknowledgment}
E. Razinkov was funded by the Russian Government support of the Program of Competitive Growth of Kazan Federal University among World's Leading Academic Centers and by Russian Foundation of Basic Research, project number 16-01-00109a. J. Matas was supported by Czech Science Foundation Project GACR P103/12/G084.




%

\end{document}